# Semi-Supervised Approach to Monitoring Clinical Depressive Symptoms in Social Media

Amir Hossein Yazdavar*, Hussein S. Al-Olimat*, Monireh Ebrahimi*, Goonmeet Bajaj*, Tanvi Banerjee*, Krishnaprasad Thirunarayan*, Jyotishman Pathak** and Amit Sheth *

* Kno.e.sis Center, Wright State University, Dayton, OH, USA

Email: {amir;hussein;monireh;goonmeet;tanvi;tkprasad;amit}@knoesis.org

** Division of Health Informatics, Cornell University, New York, NY, USA

Email: jyp2001@med.cornell.edu

*Abstract*—With the rise of social media, millions of people are routinely expressing their moods, feelings, and daily struggles with mental health issues on social media platforms like Twitter. Unlike traditional observational cohort studies conducted through questionnaires and self-reported surveys, we explore the reliable detection of clinical depression from tweets obtained unobtrusively. Based on the analysis of tweets crawled from users with self-reported depressive symptoms in their Twitter profiles, we demonstrate the potential for detecting clinical depression symptoms which emulate the PHQ-9 questionnaire clinicians use today. Our study uses a semi-supervised statistical model to evaluate how the duration of these symptoms and their expression on Twitter (in terms of word usage patterns and topical preferences) align with the medical findings reported via the PHQ-9. Our proactive and automatic screening tool is able to identify clinical depressive symptoms with an accuracy of 68% and precision of 72%.

## I. INTRODUCTION

A common global effort to manage depression involves detecting depression through survey-based methods via phone or online questionnaires[1]. However, these studies suffer from underrepresentation, sampling biases, and incomplete information. Additionally, large temporal gaps between data collection and the dissemination of findings can delay the administration of timely and appropriate remedial measures. Cognitive bias, which prevents participants from giving truthful responses, is yet another limitation [1]. In contrast, Twitter is a valuable resource for learning about users' feelings, emotions, behaviors, and decisions that reflect their mental health as they are experiencing the ups and the downs in real-time. For example, news headlines such as "Twitter fail: Teen Sent 144 Tweets Before Committing Suicide & No One Helped" and "Jim Carrey's Girlfriend: Her Last Tweet Before Committing Suicide 'Signing Off", illustrate the expression of emotional turmoils in tweets and subsequent deliberate actions in the physical world.

In recent years, much progress has been made in studying mood and mental health through social media content [2, 3, 4, 5, 6]. These studies can be categorized into two major groups; namely, lexicon-based [7, 8], and supervised [5, 9, 10, 11, 12]. These studies suggested the individual's language style, emotion, ego-network, and user engagement as discriminating features to recognize depression-indicative posts. However, the lexicon-based approaches suffer from low recall and are highly dependent on the quality of the created lexicon. On the other hand, supervised approaches require labor intensive annotation of a huge dataset. Besides, suffering from clinical depression is more than feeling down for a few days [13]. Indeed, clinical depression is diagnosed through a set of predefined symptoms which last for a fixed period.

[1]https://www.cdc.gov/mentalhealthsurveillance/

Inspired by that, we develop a statistical model which emulates traditional observational cohort studies conducted through online questionnaires by extracting, categorizing, and unobtrusively monitoring different symptoms of depression by modeling user-generated content in social media as a mixture of underlying topics evolving over time. To our knowledge, this is the first study that incorporates temporal analysis of user-generated content on social media for capturing these tell-tale symptoms. We crawled 23 million tweets posted by over 45,000 Twitter users who self-reported symptoms of depression in their profile descriptions.

The present study answers the following questions: 1) How well can textual content in social media be harnessed to reliably capture a user's symptoms of clinical depression over time and build a proactive and automatic depression screening tool? 2) Are there any underlying common themes among depressed users?

We assess the level of depression expressed in tweets for each user profile in our dataset by integrating a lexicon-based method (top-down processing) with a data-driven method (bottom-up processing). Leveraging the clinical articulation of depression, we build a depression lexicon that contains common depression symptoms from the established clinical assessment questionnaire PHQ-9 [13]. We rank the terms and compile a list of informative lexicon terms for each user and use them as seed terms to discover latent topics (depression symptoms) discussed by the subject in his/her tweets (bottom-up processing). We develop a probabilistic topic modeling over user tweets with partial supervision (by leveraging seeded clusters), named *semi-supervised topic modeling over time* (ssToT), to monitor clinical depression symptoms. We apply ssToT to derive the per user topic (depression symptoms) distribution and per topic word distribution to screen and determine a trend of symptoms over time.

The major contributions of this multidisciplinary study,

conducted by a team of computer scientists and mental health experts are two-fold: **First**, we create a lexicon of depression symptoms which are likely to appear in the generated content of depressed individuals; **second**, we develop a semi-supervised statistical model to extract, categorize, and monitor depression symptoms for continuous temporal analysis of an individual's tweets. Empirical evaluations show our model is superior to five baselines in terms of the quality of learned topics (clinical depression symptoms).

## II. RELATED WORK

Several efforts have attempted to automatically detect depression in social media content using machine learning approach. Conducting a retrospective study over the content of tweets posted by depressed and non-depressed individuals for one year, [14] characterize depression based on factors such as language, emotion, style, ego-network, and user engagement. They utilize these distinguishing characteristics to build a classifier to predict the likelihood of depression in a post [14] or in an individual [9]. In another study, [10] leverage affective aspect, linguistic style, and topics as a feature for detecting depressed communities. [15] employ various features including sentence polarity for detection of depressed individuals in Twitter. They use n-grams, psychological categories, and emoticons as features. Similarly, [16] apply a classifier for identifying potential cases of social network mental disorders. Moreover, there have been significant advances in the field by introducing a shared task [11] at the Computational Linguistics and Clinical Psychology Workshop (CLP 2015) focusing on methods for identifying depressed users on Twitter. A corpus of nearly 1,800 Twitter users was built for evaluation; among the several participants in the shared task, the best models employed topic modeling [2], and various features such as bag of words, LIWC features, metadata and clustering features [17]. Another related line of research focused on capturing suicide and self-harm signals from Twitter posts [18]. Through analysis of tweets posted by individuals attempting committing suicide, [19] indicate quantifiable signals of suicidal ideations. Moreover, the 2016 ACL Computational Linguistics and Clinical Psychology Workshop [20] defined a shared task on detecting the severity of the mental health forum posts. *All of these studies define some discriminative features to classify depression in user-generated content in a message, for a user or at a community level. However, our approach facilitates detection of depression through fine-grained temporal monitoring of subjects' behavior by analyzing the symptoms of depression mirrored in their topics of interest and word usage. Apart from that, what makes our model different is that it does not require any labeled dataset.*

In the context of *lexicon-based* approaches, [8] use a dictionary-based method for assigning an overall depression score to subjects. They count all the phrases that are matched with depression indicators without considering separate symptom categories. [21] study the usage of keyword the "depression" in tweets. They find initial evidence that individuals tweet about their depression and even disclose updates about their mental health treatment on Twitter. They found an association between excessive use of negative-emotions-related words and having a major depressive disorder. In contrast, no relation has been found in the use of positive-emotions-related words and depression. Similarly, [7] propose an NLP methodology for automatic screening of depression using a depression lexicon incorporating both metaphorical and non-metaphorical words and phrases. They perform a web search to retrieve documents containing "depression is like *" pattern. However, in natural language, words can be ambiguous. For instance, depression may be used to express different concepts such as "economic depression", "great depression", "depression era", and "tropical depression". Moreover, neurotypical people use this term to express their transient sadness. For instance, consider: "I am depressed, I have a final exam tomorrow". Furthermore, the experience of depression may be expressed implicitly, making a lexicon-based approach insufficient for accurate fine-grained analysis of depression symptoms over time. Another inherent drawback of all lexicon-based methods is their high precision at the expense of low recall and lack of context-sensitivity. For instance, "...sleep forever..." may indicate suicidal thoughts rather than the act of sleeping. In short, our study differs from existing works in that we developed a statistical model for the linguistic analysis of social media content authored by a subject by seeking depression indicators and their variation over time.

TABLE I. TWEETS SAMPLE AND THEIR ASSOCIATED SYMPTOMS

| PHQ-9 Symptoms | Short-text Document |
|---|---|
| Lack of Interest | I've not replied all day due to total lack of interest, depressed probs |
| Feeling Down | i feel like i'm falling apart. |
| Sleep Disorder | Night guys. Hope you sleep better than me. |
| Lack of Energy | so tired, so drained, so done |
| Eating Disorder | I just wanna be skinny and beautiful |
| Low Self-esteem | I am disgusted with myself. |
| Concentration Problems | I couldn't concentrate to classes at all can't stop thinking |
| Hyper/Lower Activity | so stressed out I cant do anything |
| Suicidal Thoughts | I want summer but then i don't... It'll be harder to hide my cuts. |

## III. PROPOSED APPROACH

The Diagnostic and Statistical Manual of Mental Disorders (DSM)[2] suggests that clinical depression can be diagnosed through the presence of a set of symptoms over a fixed period of time. The PHQ-9 [3] is a nine item depression scale, which incorporates DSM-V. It can be utilized to screen, diagnose, and measure the severity of depression. Our research hypothesis is that depressed individuals discuss their symptoms on Twitter. Symptoms of depression include decreased pleasure in most activities (S1), feeling down (S2), sleep disorders (S3), loss of energy (S4), a significant change in appetite (S5), feeling worthless (S6), concentration problems (S7), hyper/lower activity (S8), and suicidal thoughts (S9). This is a top-down definition of depressive disorder through its "symptomatology". To validate this hypothesis, we first manually examined symptoms in a random selection of 100 user profiles in our dataset. Table I illustrates a sample of anonymized tweets and their associated symptoms in PHQ-9. This table highlights the importance of developing appropriate models of textual content to capture depressive behavior on Twitter.

Motivated by these observations, we investigate two approaches for detecting symptoms of clinical depression on

---

Twitter, emulating the PHQ-9 questionnaire. The first approach captures clinical depression using bottom-up processing of user tweets and distributional semantics to uncover symptoms of depression via related word clusters. The second approach hybridizes the first approach with top-down processing by using the lexicon terms to guide the extraction of symptoms from tweets.

### A. Bottom-up processing:LDA

In health data mining, the problem of discovering latent topics represents a promising research area [22]. We hypothesize that by analyzing a user's topic preferences (what) and word usage (how) we can monitor depression symptoms. Our approach is based on latent variable topic models, more specifically, Latent Dirichlet Allocation (LDA). LDA is an unsupervised method that views a document as a mixture of latent topics, where a topic is a distribution of co-occuring words. Different terms expressing a related facet would be grouped together under the same topic. We apply LDA to extract latent topics discussed by users in our dataset.

Not surprisingly, the topics learned by LDA are not granular and specific enough to correspond to depressive symptoms. Several prior studies also highlight that the results from the traditional LDA do not correlate well with human judgments [23, 24]. Some work has been done to guide the discovery of latent topics in LDA by incorporating domain knowledge in different ways; from defining a set of First-Order Logic (FOL) rules [25] to constraining the occurrence of some terms together by encoding a set of Must-Links and Cannot-Links associated with the domain knowledge [26] or in the context of aspect extraction for sentiment analysis by providing some relevant terms for a few aspects [27].

The key difference between our seeding model and this study is that we supervise the topics at the token level rather than measure the distribution over a predefined list of terms. In particular, we restrict the occurrence of relevant tokens within the specified topics. We will further explain the seeding approach in the following section.

### B. Hybrid processing: Proposed ssToT Model

The basis of traditional LDA is the frequency of the co-occurrence of terms in various contexts. This syntactic approach often results in many terms from different symptom categories being merged into a single topic. By constraining symptom-related seed terms so that they only appear in a single topic, we bias the "bottom-up" learned topics to align with expected "top-down" symptom categories. In particular, we add supervision to LDA, by using terms that are strongly related to the 9 depression symptoms as seeds of the topical clusters and guide the model to aggregate semantically-related terms into the same cluster.

To generate a set of seed terms for each symptom category, we leverage the lexicon as background knowledge. In particular, in collaboration with our psychologist clinician, we built a lexicon of depression-related terms that are likely to be utilized by individuals suffering from depression. We use patient health questionnaire (PHQ-9) categories as a predefined list of depression symptoms. Furthermore, given the colloquial language of social media, we use Urban Dictionary (a crowd-sourced online dictionary of slang words and phrases)[4] for expanding the lexicon using the synset of each of the nine PHQ-9 depression symptoms categories. We also employ Big Huge Thesaurus[5] to obtain synonyms for each symptom category. The consistency of the built lexicon with psychologist's requirements has been vetted by our psychologist collaborators. After several rounds of refinement by domain experts, the final lexicon contains more than 1,620 depression-related symptoms categorized into nine different clinical depression symptom categories which are likely to appear in the tweets of individuals suffering from clinical depression.

However, there are important challenges to overcome in order to effectively leverage our lexicon for compiling a seed cluster. First, social media users often use diverse terms to express a specific concept. They use creative descriptive metaphorical phrases and explanations for symptoms. One may say, "I'm so exhausted all time" while another may say "so tired, so drained, so done" while both of these utterances discuss the unique medical concept "Lack of Energy". Second, language of social media contains polysemous words in its vocabulary. Their interpretation requires context for Word Sense Disambiguation (WSD). For instance, "Cut my finger opening a can of fruit" and "scars don't heal when you keep cutting" use "cut" in different contexts and senses.

To address the *first* challenge, our algorithm automatically generates a personalized set of seed terms per user which is a subset of the available terms in the lexicon. In this manner, a list of highly informative seeds will be generated per user. For the above examples, the term "exhausted" would be a seed for the first user while "drained" and "tired" would be the seeds for the second user. To address the *second* challenge, given the recent advances in sentiment analysis techniques [28, 29, 30], we disambiguate a polysemous word based on the sentiment polarity of its enclosing sentence. We include a term as a seed only if the enclosing context has negative sentiment. We perform sentiment analysis using the Python TextBlob[6], a standard library, which determines positive/neutral/negative polarity for any document. For the above example, "cut" is not a seed for the first user, but is a seed for the second user, as the first tweet reflects a neutral sentiment while the second tweet indicates a negative sentiment.

On the other hand, experiencing clinical depression is more than feeling down for a few days. According to PHQ-9 clinical depression symptoms should persist for a few weeks. Hence, temporal monitoring of symptoms is crucial.

## IV. ALGORITHM

Motivated by the above observations, we propose our framework to automatically analyze user behavior by continuously monitoring their social media content over time intervals. To this end, the proposed approach enriches the LDA model's expressiveness by introducing a predefined set of seed terms. We divide each user's collection of preprocessed tweets into a set of tweet buckets using a specific time interval of $d$ days. The generative process of the proposed model for a corpus $C$

---

[4] http://www.urbandictionary.com/
[5] https://words.bighugelabs.com/
[6] https://textblob.readthedocs.io/en/dev/

of individual user's tweets consisting of $B$ buckets is shown in Algorithm-1.

---

**Algorithm 1** The generative process of ssToT

---
1: **procedure** ANALYZETWITTERPROFILE
2:    **for each** symptom (topic) $s \in 1, 2, ..., 9$ **do**
3:       Draw a distribution over terms and seed sets
       $\Phi_s \sim Dirichlet(\beta)$
4:    **end for**
5:    **for each** bucket (document) $b \in 1, 2, ..., B$ **do**
6:       Draw a distribution over topics $\theta_b \sim Dirichlet(\alpha)$
7:       **for each** word $w_i \in b$ **do**
8:          Choose a symptom (topic) $s_i \sim multinomial(\theta_b)$
9:          Choose a word $w_i \sim multinomial(\Phi_{s_i})$
10:       **end for**
11:    **end for**
12: **end procedure**

---

In Algorithm-1, $\theta$ shows the distribution of symptoms over buckets while $\Phi$ is the distribution of words per symptom. We employ Gibbs Sampling to approximate the posterior distribution over the assignment of words to topics, $P(s|w)$. We then estimate $\Phi$ and $\theta$ using this posterior distribution. Our strategy for discovering symptoms (topics) differs from previous methods as we incorporate prior knowledge into the inference by assigning the pre-defined seed terms into only one of the symptoms (topics). Inspired by [23], we adapt the Gibbs Sampling equation by restricting a topic $s_i$ to a single conventional value for each user-specific seed term or phrase. Each term $w_i$ is assigned to the largest probability symptom associated with it in $\Phi$. We change the probability of a symptom over a bucket to zero if the number of seed terms associated with it is less than a threshold $\tau$. Similarly, to filter out polysemous seed terms, we aggregate the sentiment polarity of all sentences containing all seed terms over a bucket. If the aggregated polarity is positive, we assign the probability of zero to all symptoms in that bucket. Finally, we visualize the probability of each symptom over the bucket in matrix for further analysis and monitoring. Apart from that, if the probability of a symptom is more than a threshold $\tau$, the symptom would be assigned to the bucket as a label. In this manner, our model can be utilized as a multi-label classifier over a time interval. The quality of our multi-label classifier is evaluated as follows.

## V. EXPERIMENTAL RESULTS

We first discuss data collection procedure, followed by qualitative and quantitative analysis. We highlight that since this study analyzes individual's behavioral health information, which may considered as sensitive, in our datasets, we anonymized users' real identities as per the approved institutional review board (IRB) protocol.

**Dataset:** We created a dataset containing 45,000 Twitter users who self-declared their depression and 2,000 "undeclared" users who were collected randomly In particular, for collecting self-declared depressed individual's profiles, we utilize a subset of highly informative depressive indicative terms in our lexicon and find the profiles that contain these terms in their description. Afterwards, we crawled the tweets, the tweets' timestamp, and the list of friends and followers of these users. After removing the profiles with less than 100 tweets, we obtained 7,046 users with 21 million timestamped tweets, with each user contributing at most 3,200 tweets due to the Twitter Search API limitation. Next, we randomly sampled 2,000 profiles of users with self-reported depression symptoms and 2,000 random users who do not have any depression terms in their profile descriptions. We denote this subset of 4,000 users by $U$. We preprocess these tweets by changing the space delimiter into underscore so all phrases in the tweets that are listed in our lexicon as a seed phrase ( e.g., lack_of_interest). Topic modeling is a word-level approach while most of the depression seeds in our lexicon are phrases. Consequently, seed phrase replacement plays an integral role in the success of our algorithm. Next, we apply platform-specific filtering, followed by non-ASCII character and stopword removal, as well as lemmatization. Platform-specific filtering includes substituting retweets ("RT @username" by RT), user mentions ("@username" by MENTION), and hyperlinks (by URL). For spelling correction, we utilize the PyEnchant spell checker library[7]. Furthermore, alphabetic character repetition (writing identical characters in sequence for emphasis, e.g., fatttttttt, sleeeeeep) is addressed by defining regular expressions and enhancing the available NLTK tweet tokenizer library[8].

### A. Qualitative Results

*1) Discovery of depressive symptoms:* Our ssToT model discovers depressive symptoms as latent topics from sliding window on buckets of timestamped tweets posted by users. We rank the top terms in each symptom $p(w|s)$ in descending order. Table II illustrates the sample of topics learned by ssToT and LDA model. The seeded words for the ssToT model are boldfaced, and words that are judged as relevant are italicized. We observe that by constraining seed terms to a specific symptom, the discovered terms are more relevant to that category. For example, in LDA model Topic 8 contains three terms relevant to "Sleep Disorder" (S3); however, it also contains lots of irrelevant terms which makes the emphasis of this topic off-target. Although Topic 6 from LDA contains terms relevant to "Eating Disorder" (S5), it also contains some terms related to "Sleep Disorder" and "Suicidal Thoughts" (S9). Similarly, for Topic 3, it contains terms associated with both the "Eating Disorder" and "Suicidal Thoughts" categories. Therefore, the topics discovered with LDA are not interpretable for the purpose of this study.

In contrast, the topics learned from the ssToT model contain more relevant terms associated with symptom category and more interpretable topics (see Table II). Additionally, the ssToT model also captures acronyms that people use in social media; for instance, in symptom 5 (Eating Disorder) "ugw" stands for "Ultimate Goal Weight" and "mfp" for "More Food Please", or in symptom 2 (Lack of Interest) "idec" for "I Don't Even Care". We also observe the excessive usage of expressive interjections in language used by depressed users. Terms such as "argh" (showing frustration), "aw" (indicative of disappointment), "feh" (indicative of feeling underwhelmed), "ew" (denoting disgust), "Huh" (indicator of confusion), "phew" (showing relief) were mostly discovered in their related symptoms category.

Furthermore, we observe that there are common themes and triggers of clinical depression at the community level

---

| Model | Label | Top Words | UMass | UCI | NPMI |
|---|---|---|---|---|---|
| ssToT | Sleep Disorder | *cant sleep*, **wanna sleep**, *night*, *nighttime*, *sleepy*, **need to sleep**, *hour*, **sleepover**, **bedtime**, **go to sleep**, mess, *dream*, *midnight crying*, painful, *5:00 AM*, guilt, struggle, *headaches tonight*, *morning*, *coffee*, *duvet*, *hungover*, bbe | −1.23 | −1.28 | −0.01 |
| | Eating Disorder | **fat**, *eat*, *kg*, *weight loss*, *negative calories*, *lbs*, *thin*, *my thighs*, *paper thin*, *binge*, **eating disorder**, *abs*, *stomach*, **bulimic**, hating, *salad*, pretend, *gain*, **starve**, *mcdonalds*, *bones*, *chubby*, flat, skip, *wears*, *kcal*, **puffy**, *hippo*, *mfp*, *ugw* | −1.18 | 0.20 | 0.02 |
| | Suicidal Thoughts | **self harm**, *cut*, **suicide**, *live*, *scar*, *blade*, *dead*, *alive*, *bleed*, *need my blade*, **death**, *hanging*, *deserve pain*, *kill me now*, *gun*, **want to die**, *knife*, daisies, opinion, meh, *razor*, *sharp*, *wrists*, pictures, *never wake up*, **wanna cut**, stfu, *ew* | −0.66 | 1.13 | 0.09 |
| LDA | Topic 8 | *Sleepover*, september, lost, interest, exaggerating, its my fault, ugh, *sleep*, skin, dish, saved, *wake up*, blocked, blow, ipad, touches | −1.44 | −2.84 | −0.1 |
| | Topic 6 | *thigh*, blood, *big*, *beautiful*, *thin*, smile, sleep, blood, leave, stay, worthless, *fat*, tear, pretending, sadness, fake, ugly, god, skin, eat, morning | −3.31 | −2.65 | −0.08 |
| | Topic 3 | *Blade*, ugly, fat, *blood*, smile, mirror, call, fit, eat, stay, beautiful, sleep, big, tear, sad, devil, god, skin, music | −2.69 | −3.69 | −0.09 |

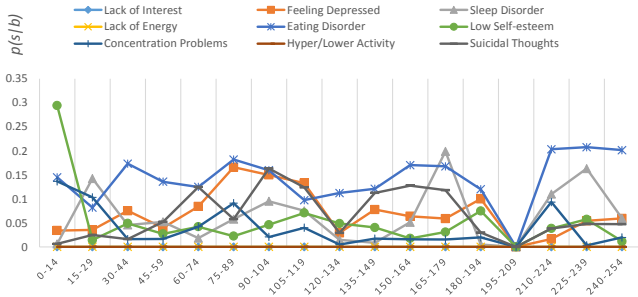

Fig. 1. Sample trend analysis of depressive symptoms

that they do not existing in PHQ-9. In most cases, depressed users discuss their family and friend problems and the need for their support. For example the topic {family, hugs, attention, parents, competition, daddy, mums, sigh, grandma, losing, maam, friendless, love, friend, mommy, people, boyf, gf} shows that the person is suffering from a relationship problem. Another common theme is school and academic stress {schools, college, exam, classmate, friendless, teacher, assignment}.

To visualize depressive symptoms discovered for each user over a specific period, we keep the topics which contain at least certain number of seed terms as dominant words (among the top 20 terms associated with that topic) and discard the others.

Figure 1 depicts the sample of various depressive symptoms learned by ssToT and their distribution over time. It shows *when* and for *how long* a specific depressive symptom *occurred*. To check the validity of each topical trend, human annotators were asked to manually annotate all the buckets for the presence of all symptoms. Further details are provided in the next section.

## VI. QUANTITATIVE RESULTS

Since ssToT is based on semi-supervised learning, it can be considered a clustering based approach and evaluated based on clustering evaluation measures. In addition, we are also able to employ classification accuracy to evaluate the performance of our method for symptom discovery. In this section, we first discuss the state of the art in clustering measures and how we adopt these measures to compare the performance of the ssToT model with existing methods. Then, we discuss the process of creating a ground truth dataset for evaluating the performance of the ssToT for discovering depressive symptoms as a multi-label classifier over different time periods.

**Coherence Measures:** Topic coherence measures score a single topic by identifying the degree of semantic similarity between high-scoring words in that topic. In this manner, we can distinguish between semantically interpretable topics and those which are artifacts of statistical inference. The state of the art for this evaluation criterion can be grouped into two major categories: intrinsic and extrinsic measures. Intrinsic measures evaluate the amount of information encoded by the topics over the original corpus used to train the topic models.

Another common intrinsic measure is UMass presented by [31], which measures the word co-occurrence in documents:

$$UMass(w_i, w_j) = log\frac{D(w_i, w_j) + \epsilon}{D(w_i)}$$

where $D(w_i, w_j)$ counts the number of documents containing both $w_i$ and $w_j$ words, and $D(w_i)$ counts the ones containing $w_i$, over the same training corpus, and $\epsilon$ is the smoothing factor. The UMass metric computes these probabilities over the same corpus used to train the topic models.

Conversely, extrinsic evaluation metrics estimate the word co-occurrence statistics on external datasets such as Wikipedia. The UCI metric introduced by [32] utilizes the Pointwise Mutual Information (PMI) between two words,

$$UCI(w_i, w_j) = log\frac{p(w_i, w_j) + \epsilon}{p(w_i)p(w_j)}$$

where the word probabilities are calculated by counting word co-occurrence in a sliding window over an external dataset such as Wikipedia. Recently, another topic coherence measurement has been introduced by [33] which considers context vectors for every topic's top word. For every word w, a context vector is generated using word co-occurrence counts employing context window of size +-n surrounding that word. By calculating Normalized PMI (NPMI), they showed their method has a strong correlation with human topic coherence rating. The higher the topic coherence measure score, the higher the quality of the topics. This, in turn, leads to better topic interpretability, given that our purpose is to extract meaningful and interpretable topics associable with depressive symptoms. We used Palmetto[9] for measuring the quality of topics learned based on NMPI and UCI measures (Wikipedia as an external corpus). UMass was measured by creating our Lucene index on tweets from users in set U (intrinsic evaluation)[10]. Table II shows a sample of the coherency of topics learned (symptoms) for LDA and ssToT models based on UMass, UCI, and NPMI metrics. We can clearly see that the topics learned by the ssToT model are more coherent for all the three measures.

**Baselines:** To further evaluate the ssToT-learned topics, we compare them with the topics obtained from a set of existing unsupervised and semi-supervised approaches. **k-means:** A clustering approach based on distributional similarity employing cosine similarity measure. **LSA:** An unsupervised approach that gleans distributional semantics by clustering correlated terms into latent topics using singular value decomposition. **LDA:** A Bayesian approach that represents a document as a mixture of topics. **BTM:** A state-of-the-art unsupervised topic modeling framework for short texts which utilizes distributed representations of words and phrases [34]. **Partially Labeled LDA:** A semi-supervised topic model which constrains latent topics to align them with human-provided labels [24].

To determine the number of topics for all LDA variants, we use perplexity using 80% of the data to train and 20% to test. We choose 15 topics as a proper level of granularity as it has the lowest perplexity and is suitable for our task. We set the number of Gibbs iterations to 1,000, $\alpha$ to 0.5, $\beta$ to 0.1 and the rest of the parameters to default values. We use the Stanford Topic Modeling Toolbox[11] to run all LDA variants except BTM, which is downloaded from its author's webpage[12]. We use cosine similarity as a distance function for k-means. Table III denotes the average coherence score for each model. Due to space limitations, we only report the average coherence of all symptoms for each algorithm. Coherence measures judge each model's output based on how well they represent a specific topic. This aligns with our objective of providing outputs that are well associated with depressive symptoms rather than some generic set of terms grouped together. These numbers indicate that the ssToT model outperforms other state-of-the-art techniques regardless of the corpus that probabilities are gained from: Wikipedia for UCI or the same corpus in UMass. We note that although, on average, the ssToT model outperforms the other five baselines in terms of discovering coherent topics, there are rare exceptions. For instance, the topic containing {fat, time, feel, dinner, weight, eat, hate, skinny} learned by the BTM algorithm about Eating Disorder has scores of -0.19, 0.8, and 0.08 for UMass, UCI, and NPMI respectively, which implies that it is more coherent compared to its associated topics learned by ssToT (see Table III). However, when we further analyzed the rest of the topics learned by BTM, we noticed the poor quality of the other learned topics, as the average coherency also indicates.

TABLE III. Average coherency of different models vs. ssToT

| Model | UMass | UCI | NPMI |
|-------|-------|-----|------|
| LDA | -2.68 | -3.03 | -0.109 |
| BTM | -1.42 | -2.18 | -0.058 |
| P-LDA | -1.12 | -2.48 | -0.123 |
| K-Mean | -1.70 | -2.95 | -0.102 |
| LSA | -1.43 | -3.23 | -0.107 |
| **ssToT** | -1.00 | -1.62 | -0.026 |

Fig. 2. Symptom distribution in the gold standard dataset

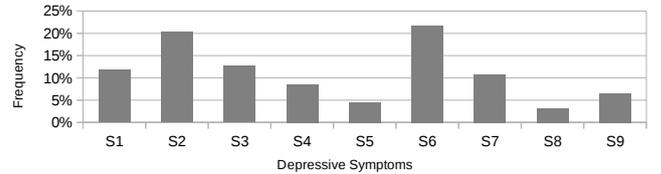

### VII. Symptom Prediction (Multi-label Classification)

We showed how ssToT is highly effective in terms of the quality of learned topics (depressive symptoms). In this section, we further investigate the power of the ssToT model as a multi-label classifier. Specifically, we try to predict the correct set of labels (depressive symptoms) for each bucket of tweets. We build a ground truth dataset of 10400 tweets in 192 buckets. Each bucket contains tweets that are posted by the user within span of 14 days (in compliance with PHQ-9). Tweets are selected from a randomly sampled subset of both self-reported depressed users and random users. Three human judges (undergraduate students who are native English speakers) manually annotated each tweet using the nine PHQ-9 categories as labels. Additionally, the non-relevant tweets that do not show any depressive symptoms, have been labeled as "cannot judge". The average inter-annotator agreement is K=0.74 based on Cohen's Kappa statistics. We build a labeled bucket by merging the labeled tweets. Figure 2 depicts symptoms distribution in the gold standard dataset.

Our semi-supervised ssToT model does not use the labeled data during training. In particular, we are not supervising the LDA model with a labeled-dataset like Labeled-LDA [35]. Instead, we are using the labeled dataset for evaluating the performance of the ssToT model in assigning a set of symptoms to each bucket. We evaluate the performance of the ssToT in terms of the average precision, recall, and F-score in detecting depressive symptoms for each bucket of tweets when tested against human judgment. We show that our ssToT model can predict the *presence* of each of the nine depressive symptoms for a different bucket with an accuracy of 0.68 (see Table IV) which gives a precision of 0.72 *on average*.

We observed that the best results were achieved for "Lack of Interest" (symptom 1) with a 0.90 F-Measure and the worst result was obtained for "Concentration Problems" (symptom 7). We noticed that our ssToT model works well with less descriptive symptoms since it generates relevant seed terms to discover the latent symptom. For instance, consider the following tweets selected from our dataset: "Overthinking always destroy my mood", "This essay is dragging so much, can't deal with essays and revisions any more :(", "I need a break from my thoughts", and "my head is such a mess right now". The first tweet contains the depression-indicative keywords "overthinking", which can easily be interpreted as "Concentration Problems" while the other utterances, although they are labeled into the same symptom category (according to our human annotator), are descriptive and metaphoric and do not contain any depressive-indicative term. However, there are some tweets such as "overthinking killed my happiness" that, even though they contain the depression-indicative terms of "Concentration Problems", (overthinking in this case) cannot be grouped into "Concentration Problems" category. Such examples contribute to a high number of false positives and low precision for this category. Furthermore, sometimes correctly determining the category of depressive symptoms is challenging even for a human. For instance, in the tweet "Need to sleep, always so f***ing tired" one may categorize it as "Lack of Energy" while another may consider it as "Sleep Disorder".

To further test the robustness of our ssToT model as a multi-label classifier, we compare its results to common supervised approaches for performing multi-label classification, namely the *binary relevance* (BR) and *classifier chains* (CC) methods [36]. The BR method transforms the problem of multi-label into multiple binary models by creating one model for each label. For this, each binary model will be trained to predict the relevance of each of the labels. On the other hand, CC is a chaining method that uses L binary transformations (one for each label) similar to BR, but it can also model label correlations while maintaining acceptable computational complexity. As supervised baseline approaches, Multinomial Naive Bayes and SVM models have been chosen for the two aforementioned methods. These two models have been widely utilized as a baseline by most previous studies [24]. Note that in the task of supervised multi-label classification, labels are available during training. We used Meka (a Multi-label Extension to WEKA)[13] for building the baselines. We use a bag-of-word model and perform 10-fold cross-validation to evaluate accuracy for each symptom (see Table IV).

These results show that in spite of the semi-supervised nature of ssToT model, it is competitive with supervised approaches and improves upon them in five out of nine symptom classification in terms of F-score along with providing better averaged accuracy. A key advantage of ssToT over supervised approach is that it does not require labor-intensive, expensive, and time-consuming manual annotation of data in training.

Our study has limitations. For users who do not generate ample content on their profiles or are reluctant to publicly reveal their depressive symptoms, we cannot assess their depressive behaviors. Additionally, we only detect the presence, duration, and frequency of symptoms rather than their severity. Furthermore, more severely depressed individuals may be more inclined to publicly express their depression and biasing our sample.

## VIII. Conclusion and Future work

We demonstrated the impact of social media on extraction and timely monitoring of depression symptoms. We developed a statistical model using a hybrid approach that combines a lexicon-based technique with a semi-supervised topic modeling technique to extract per user topic distribution (clinical,

---

---

TABLE IV. Model's performance for bucket level symptom prediction, (P:Precision, R:Recall, F:F-Score, AA. is the average accuracy for each model.)

| Model | AA. | | S1 | S2 | S3 | S4 | S5 | S6 | S7 | S8 | S9 |
|---|---|---|---|---|---|---|---|---|---|---|---|---|
| BR-MNB | 0.66 | P | 0.68 | 0.94 | 0.67 | 0.62 | 0.71 | 0.96 | 0.63 | 0.38 | 0.96 |
| | | R | 0.74 | 0.78 | 0.72 | 0.59 | 0.86 | 0.95 | 0.82 | 0.86 | 0.93 |
| | | F | 0.74 | 0.81 | 0.70 | 0.60 | 0.89 | 0.90 | 0.73 | 0.84 | **0.96** |
| CC-MNB | 0.63 | P | 0.73 | 0.95 | 0.73 | 0.61 | 0.90 | 0.99 | 0.70 | 0.71 | 0.91 |
| | | R | 0.66 | 0.80 | 0.68 | 0.55 | 0.84 | 0.87 | 0.65 | 0.85 | 0.80 |
| | | F | 0.72 | 0.83 | 0.73 | 0.57 | 0.91 | 0.92 | 0.72 | **0.90** | 0.88 |
| BR-SVM | 0.695 | P | 0.88 | 0.81 | 0.91 | 0.94 | 0.98 | 0.74 | 0.62 | 0.80 | 0.79 |
| | | R | 0.69 | 0.94 | 0.69 | 0.42 | 0.84 | 0.96 | 0.80 | 0.80 | 0.82 |
| | | F | 0.79 | 0.85 | 0.76 | 0.53 | 0.91 | 0.93 | 0.80 | 0.88 | 0.90 |
| CC-SVM | **0.71** | P | 0.87 | 0.93 | 0.79 | 0.93 | 0.97 | 0.98 | 0.74 | 0.92 | 1.0 |
| | | R | 0.70 | 0.93 | 0.66 | 0.40 | 0.83 | 0.94 | 0.79 | 0.80 | 0.84 |
| | | F | 0.75 | 0.86 | 0.75 | 0.51 | 0.91 | **0.95** | **0.82** | 0.89 | 0.91 |
| ssToT | 0.68 | P | 0.87 | 0.93 | 0.79 | 0.91 | 0.97 | 0.98 | 0.74 | 0.92 | 0.79 |
| | | R | 0.93 | 0.98 | 0.69 | 0.53 | 0.90 | 0.92 | 0.19 | 0.34 | 0.68 |
| | | F | **0.90** | **0.89** | **0.78** | **0.68** | **0.93** | 0.82 | 0.30 | 0.38 | 0.68 |

symptomatic of depression) and per topic word distribution (symptom indicators) by textual analysis of tweets over different time windows. Our approach complements the current questionnaire-driven diagnostic tools by gleaning depression symptoms in a continuous and unobtrusive manner. Our experimental results reveal that there are significant differences in the topic preferences and word usage pattern of the self-declared depressed group from random users in our dataset which indicates the competency of our model for this task. Our model yields promising results with an accuracy of 68% and a precision of 72% for capturing depression symptoms per user over a time interval which is competitive with a fully supervised approach. In future, we plan to apply our approach to various data sources such as longitudinal electronic health record (EHR) systems and private insurance reimbursement and claims data, to develop a robust "big data" platform for detecting clinical depressive behavior at the community level.

## IX. Acknowledgement

We are thankful to Surendra Marupudi and Ankita Saxena for helping us with data collection. We also thank Jibril Ikharo for his proofreading. Research reported in this publication was supported in part by NIMH of the National Institutes of Health (NIH) under award number R01MH105384-01A1.